\documentclass[letterpaper, 10 pt, conference]{ieeeconf}  % Comment this line out if you need a4paper

\IEEEoverridecommandlockouts                              % This command is only needed if 
\usepackage{algorithm}
\usepackage{algpseudocode}
\usepackage{array}
\usepackage[caption=false,font=small,labelfont=sf,textfont=sf]{subfig}
\usepackage{textcomp}
\usepackage{stfloats}
\usepackage{url}
\usepackage{verbatim}
\usepackage{graphicx}
\usepackage[misc]{ifsym} 
\usepackage{subfloat}
\usepackage{times}
\usepackage{amsmath,amssymb,amsopn,amstext,amsfonts}
\usepackage{cancel}
\usepackage[space]{cite}
\usepackage{pdfsync}
\usepackage{balance}
\usepackage{color}
\usepackage{mathtools}
\usepackage{bm}

\usepackage{diagbox}
\usepackage{float}
\usepackage{epstopdf}
\usepackage{pifont}
\usepackage{fixltx2e}
\usepackage{amsmath}
\usepackage{multirow}
\usepackage{booktabs}
\usepackage{svg}
\usepackage{threeparttable}
\usepackage{caption}
\usepackage{lipsum}
\usepackage[linkcolor=black,citecolor=black,urlcolor=black,colorlinks=true]{hyperref}
\usepackage{makecell}

\newcommand{\envpanel}[2]{%
    \begin{minipage}[t]{0.32\textwidth}
        \centering
        \includegraphics[width=\linewidth]{#1}
        \par\vspace{2pt}
        {\small #2}
    \end{minipage}%
}

\begin{document}
\captionsetup{font={small}}
\title{\LARGE \bf
GR2PO: Group Relative Return Policy Optimization for Continuous Robot Control
}
\author{Pengqin Wang, Qiming Zhang, Shaojie Shen, and Jun Ma$^{\textrm{\Letter}}$% <-this % stops a space
\thanks{Pengqin Wang and Jun Ma are with the Robotics and Autonomous Systems Thrust, The Hong Kong University of Science and Technology (Guangzhou), Guangzhou, China, also with the Division of Emerging Interdisciplinary Areas, the Hong Kong University of Science and Technology, Hong Kong SAR, China (pwangas@connect.ust.hk; jun.ma@ust.hk).
        {\tt\small }}%
\thanks{Qiming Zhang is with the Intelligent Transportation Thrust, The Hong Kong University of Science and Technology (Guangzhou), Guangzhou, China (qzhang255@connect.hkust-gz.edu.cn).
        {\tt\small }}%
\thanks{Shaojie Shen is with the Department of Electronic and Computer Engineering, the Hong Kong University of Science and Technology, Hong Kong SAR, China (eeshaojie@ust.hk).
        {\tt\small }}%
\thanks{{\Letter} Corresponding author}
}

\maketitle
\pagestyle{empty}  % no page number for the second and the later pages
\thispagestyle{empty} % no page number for the first page

%%%%%%%%%%%%%%%%%%%%%%%%%%%%%%%%%%%%%%%%%%%%%%%%%%%%%%%%%%%%%%%%%%%%%%%%%%%%%%%%
\begin{abstract}

Actor-critic architecture has been widely used in continuous robot control. However, they rely on learning a value network, introducing additional computational overhead during training. Moreover, policy learning may also be affected by the approximation error of value estimation. Critic-free group relative policy optimization methods provide a simpler training approach by removing the need for a critic. However, they fail to learn long-term action outcomes when directly applying immediate rewards to policy optimization in dense-reward environments. To address these problems, we propose Group Relative Return Policy Optimization (GR2PO), a critic-free reinforcement learning framework for continuous robot control. GR2PO estimates the discounted returns from the parallelly collected trajectories, performs group normalization at each rollout time index, and uses relative advantages and clipped targets to update the policy. To evaluate the effectiveness of the proposed framework, we instantiate it on robot control simulation environments and deploy the model to a real-world edge device. The results show that GR2PO significantly outperforms critic-free baselines that use immediate rewards and performs competitively against state-of-the-art actor-critic methods. Furthermore, GR2PO demonstrates competitive training efficiency. Inference tests on NVIDIA Jetson TX2 demonstrate the feasibility of deploying the learned policies on edge platforms. Further ablation experiments analyze the effects of parallel group size, return estimation methods, and target clipping ratio on learning performance. To support follow-up research, we will make the complete code publicly available after the paper is accepted, including the framework implementation, experimental configuration, and training and evaluation scripts.

\end{abstract}

%%%%%%%%%%%%%%%%%%%%%%%%%%%%%%%%%%%%%%%%%%%%%%%%%%%%%%%%%%%%%%%%%%%%%%%%%%%%%%%%
\section{Introduction}

Continuous robot control tasks require the policy to make decisions in a continuous action space to achieve strong long-term performance\cite{dreamwaq,extremeparkour,dribblebot,soccershooting}. Actor-critic architecture has been widely applied to these tasks by learning value functions to support policy optimization~\cite{peters2008, ddpg, ppo, sac}. However, it requires consistent training of the value network in addition to policy learning, which increases training costs\cite{rloo}. More importantly, the policy learning may also be affected by the approximation error of value estimation\cite{td3, ilyas2020}. This makes it necessary for us to explore a more concise alternative. In this case, the key challenge is to effectively leverage long-term behavior outcomes without value network learning.

\begin{figure}[!t]
    \centering
    \includegraphics[width=\linewidth]{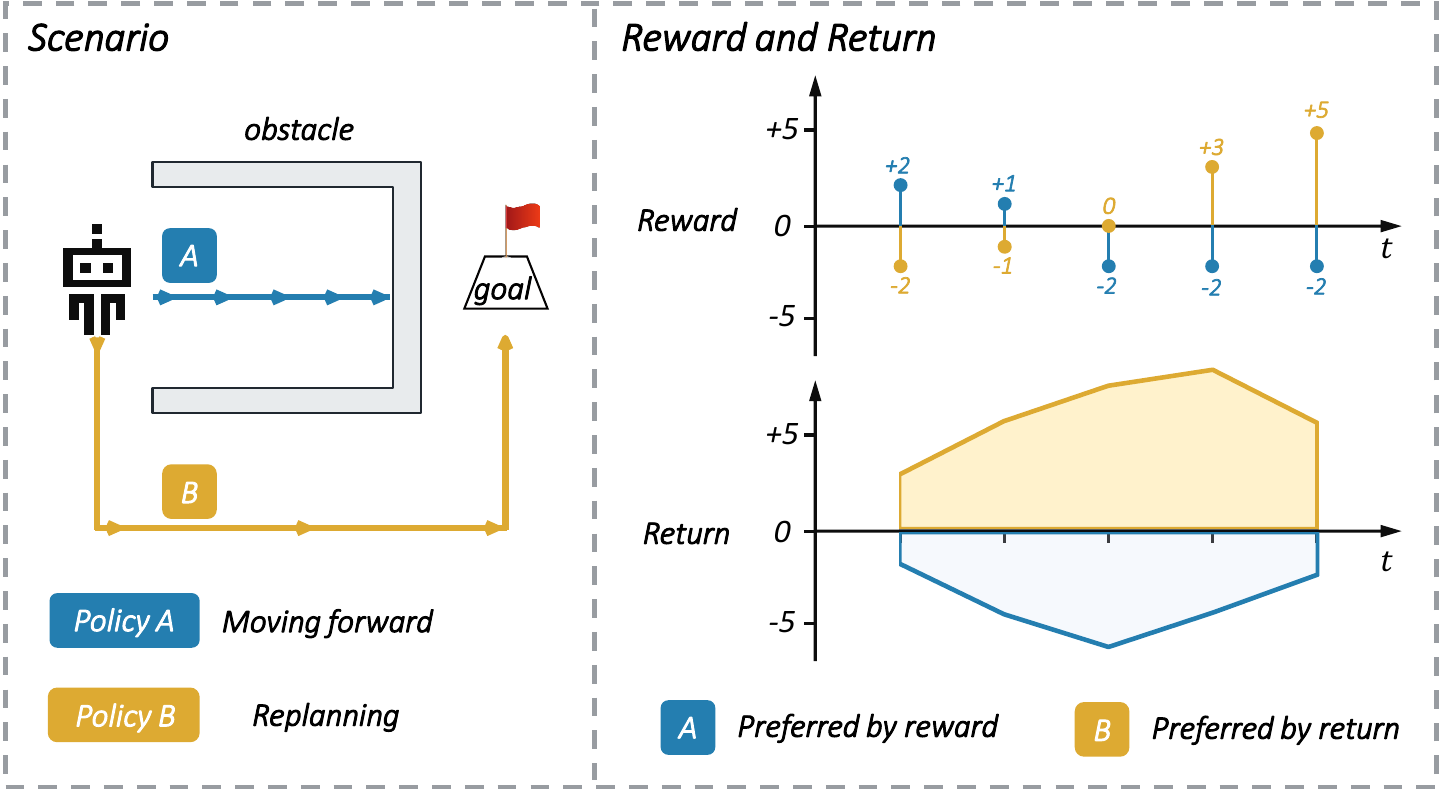}
    \caption{
        Motivation for incorporating return into group relative advantage estimation. In this robot navigation illustration, reward-oriented and return-oriented lead to different policy learning results. Evaluating actions based on immediate rewards may lead to a preference for short-term gains. The return-oriented evaluation takes future benefits into consideration, learning effective policies.
    }
    \label{fig:motivation}
    \vspace{-2.0em}
\end{figure}

The success of group relative policy optimization methods~\cite{grpo} in large language model (LLM) has shown the potential to update the policy based on the relative performance of samples within a group, avoiding the need to learn separate critics. However, complete response ratings in language generation have different temporal implications than single-step rewards in robot control. In a dense reward environment, immediate rewards describe only the local effect of actions and cannot represent future feedback. In this case, higher current rewards do not necessarily correspond to better long-term results. When the learning signal is based solely on immediate rewards, the actor policy will be driven toward short-term advantage. As shown in Fig.~\ref{fig:motivation}, moving directly towards the goal leads to high short-term rewards, but cannot reach the goal due to obstacles. In contrast, the replanning action may temporarily reduce immediate rewards but create conditions for final success. Therefore, it remains important to incorporate future benefits into the group relative evaluation~\cite{grpo_classical}.

To address these problems, we propose Group Relative Return Policy Optimization (GR2PO), which combines temporal information within a rollout with relative comparisons across rollouts. As shown in Fig.~\ref{fig:gr2po_framework}, GR2PO first collects environmental interaction trajectories in parallel, estimates returns through reverse discounted accumulation, and accounts for episode boundaries in the calculation. Then, the returns of the parallel groups at each rollout time index are normalized within groups to construct the relative advantages. Finally, the relative advantages and clipped targets are used to update the policy. Return estimation brings future outcomes, while group normalization provides relative evaluations without learning-based value functions. Our framework introduces long-term returns into group relative policy optimization for dense-reward control, removing the requirement to train a critic.

We summarize the contributions of this paper as follows.

\begin{itemize}
	\item 
    We propose GR2PO, a critic-free reinforcement learning framework for dense-reward continuous robot control. GR2PO incorporates future outcomes into group relative policy optimization by return estimation, without learning a value function.

    \item 
    We design a relative advantage construction mechanism, removing the requirements for samples with a group to share the same state. It organizes cross-environment return comparisons, normalizing returns at each time index across parallel trajectories to construct relative advantages. This directly integrates group relative advantage estimation with parallel sampling, without requiring multiple branching rollouts from the same state.

    \item
    We evaluate the effectiveness of the proposed framework on continuous control simulation experiments and deploy the model to real-world edge devices. The results show that GR2PO significantly improves learning performance over the immediate reward normalization baseline. Moreover, our framework achieves competitive returns and training costs compared with state-of-the-art actor-critic approaches.

    \item
    We systematically analyzed the key design factors that affect the performance of GR2PO. Through ablation experiments on parallel group size, return estimation methods, and target clipping ratio, we verify the advantages of Monte Carlo return estimation and moderate clipping. Furthermore, we show the performance improvement trend when increasing the group size and adjusting the training configuration under a fixed total sampling budget.

\end{itemize}

\section{Related Work}

\subsection{Actor–Critic Methods for Continuous Control}

Actor-critic methods serve as important technologies for continuous robot control, combining policy optimization and value estimation~\cite{peters2008}. DDPG~\cite{ddpg} combines deterministic policy gradients with value function approximations and achieves off-policy learning through experience replay and target networks. Based on the above, TD3~\cite{td3} addresses value-function overestimation from approximation error by using dual value estimation, delayed policy update, and target policy smoothing, improving training stability. PPO~\cite{ppo} uses clipped targets to limit policy optimization within a reasonable range and supports multiple rounds of mini-batch optimization on collected trajectories. SAC~\cite{sac} optimizes the policy and the value function under the framework of maximum entropy. Based on the reuse of past experiences and double Q-learning, it has demonstrated excellent sample efficiency and training performance in continuous control tasks.

The above actor-critic approaches have advanced continuous control from different perspectives, including policy update, value estimation, and data utilization. However, they rely on learning a value network, introducing additional computational overhead during training. Furthermore, the actor performance largely depends on the quality of the learned critic. In this paper, we study critic-free policy optimization by constructing the advantages through return estimation in parallel rollouts and group relative comparisons, while retaining the clipping policy update mechanism. We select PPO~\cite{ppo} and SAC~\cite{sac} as the representative on-policy and off-policy baselines to evaluate the control performance and training computational efficiency of our framework.

\subsection{Critic-Free Policy Optimization}

Critic-free policy optimization has a long research foundation. REINFORCE~\cite{reinforce} directly constructs the policy gradient using sampling rewards, providing a foundation for policy learning based on trajectory feedback. The success of GRPO~\cite{grpo} in reinforcement learning for LLM has promoted the development of group relative policy optimization. This method constructs relative advantages based on the sample scores within the group, avoiding the need to train separate value models. The result supervision uses scores from complete responses, while the process supervision accumulates normalized rewards.

Recent studies have further strengthened critic-free policy optimization~\cite{drgrpo}. REINFORCE++~\cite{reinforcepp} integrates global advantage normalization, clipped targets and policy update for language model alignment. GiGPO~\cite{gigpo} achieves fine-grained credit assignment for LLM agents using trajectory-level evaluation with action comparison under the same conditions. Group Policy Gradient~\cite{gpg} uses Monte Carlo returns and group baselines to calculate advantages, and supports grouping by state or time. A-GRPO~\cite{agrpo} adopts time-step normalization, trajectory-level ranking, and a Lagrangian mechanism to handle control tasks with terminal constraints. These works \cite{gpg,agrpo} are closely related to our advantage construction through the use of time-indexed group baselines and return normalization.

This paper focuses on dense-reward continuous robot control. The effects of immediate rewards and discounted returns as group relative learning signals are directly evaluated on simulation environments.
Through the analysis of parallel group size, reward estimation methods and clipping ratio parameters, this paper further studies the practical performance of this method and its sensitivity to key design choices.

\section{Methodology}

\begin{figure*}[!t]
    \vspace{1.0em}
    \centering
    \includegraphics[width=\textwidth]{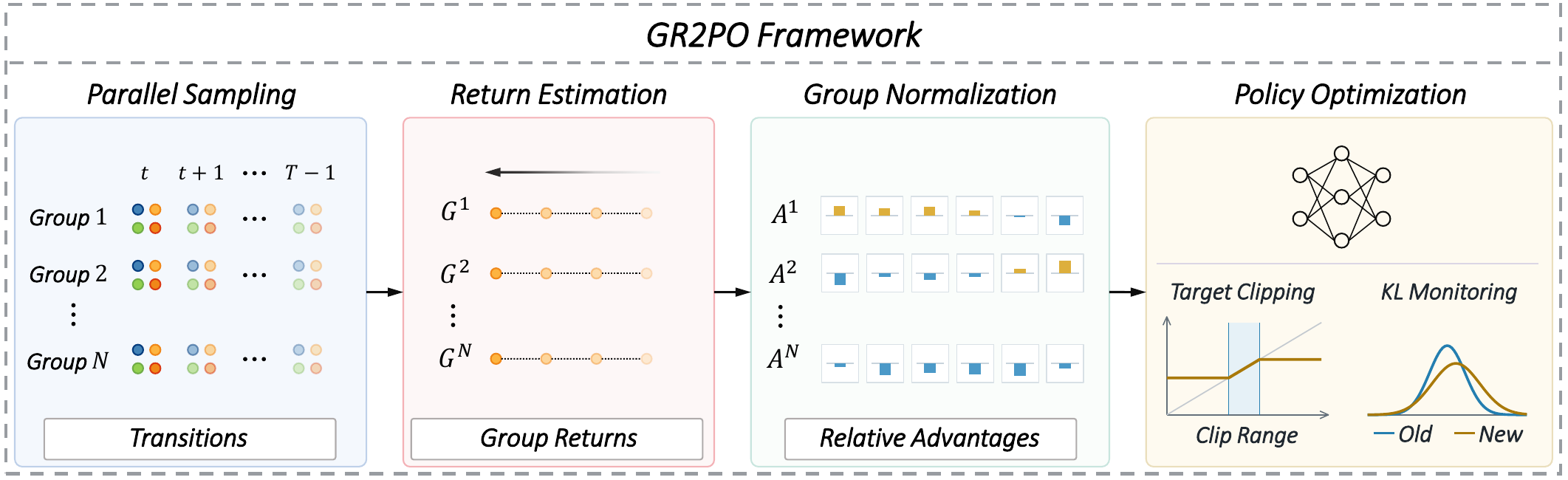}
    \caption{
        Overview of GR2PO framework. 
        Group rollouts are collected from parallel environments. Then the group returns are estimated through reverse discounted accumulation. Returns are normalized across groups at each time index to compute relative advantages, which guide policy optimization through a clipped surrogate objective with entropy regularization and KL-based early stopping. The updated actor is used to collect rollouts. GR2PO requires no learned critic.
    }
    \label{fig:gr2po_framework}
\end{figure*}

We propose Group Relative Return Policy Optimization (GR2PO), a critic-free policy optimization framework for continuous robot control, as shown in Fig.~\ref{fig:gr2po_framework}. GR2PO first collects parallel rollouts, estimates returns via reverse discounted accumulation, then performs group normalization to compute relative advantages, and uses clipped targets to update the policy. The overall training pipeline removes the requirements for value function learning.

\subsection{Problem Formulation}

We model the continuous robot control as a Markov Decision Process (MDP). The state and action are respectively denoted as $s_t\in\mathcal S$ and $a_t\in\mathcal A$, and the action space $\mathcal A$ is continuous. At timestep $t$, the robot selects actions according to the policy $\pi_\theta(a_t\mid s_t)$, receives the reward $r_t$ and moves to next state $s_{t+1}$. The learning objective is to maximize the expected discounted returns:
\begin{equation}
J(\theta)=
\mathbb{E}_{\tau\sim\pi_\theta}
\left[
\sum_{t=0}^{H_\tau-1}\gamma^t r_t
\right],
\label{eq:objective}
\end{equation}
where $\gamma\in[0,1)$ is the discount factor and $H_\tau$ represents the length of the episodes.

\subsection{Parallel Rollout Collection}

The policy $\pi_{\theta_{\mathrm{old}}}$ is performed during each sampling iteration. The training data consists of interaction rollouts with length $T$ collected from $N$ parallel environments:
\begin{equation}
\mathcal{D}=
\left\{
\left(
s_t^{(i)}, a_t^{(i)}, r_t^{(i)}, d_t^{(i)}
\right)
\right\}_{i=1,\ldots,N;\,t=0,\ldots,T-1},
\label{eq:rollout_data}
\end{equation}
where $i$ is the environment index, and $d_t^{(i)}$ shows whether the episodes is terminated after the transition. The log probability of the sampled action under the old policy is saved for policy update.
We define the number of parallel environments $N$ as the group size. Different environments perform state transitions independently. Samples from different environments with the same time index are considered as a group for normalization. In this case, we remove the requirement that all samples within a group share the same state.

\subsection{Group Relative Advantage Estimation}

In dense-reward control tasks, the immediate reward can only reflect the local effect of actions, making long-term outcomes unavailable. Therefore, it becomes important to evaluate the future returns. In our work, we perform return estimation through reverse discounted accumulation:
\begin{equation}
\begin{aligned}
G_t^{(i)}
&=r_t^{(i)}
+\gamma\bigl(1-d_t^{(i)}\bigr)G_{t+1}^{(i)},\\
G_{T-1}^{(i)}&=0.
\end{aligned}
\label{eq:return_estimation}
\end{equation}

When the episode is terminated, the accumulation will be blocked using a mask, preventing the reset reward from being included in the previous episode. For samples that are not yet finished but reach the rollout boundary, zero-tailed values are used and no value function bootstrapping is performed.

The mean and the sample standard deviation of the returns at each rollout time index $t$ are calculated:
\begin{equation}
\mu_t=\frac{1}{N}\sum_{i=1}^{N}G_t^{(i)},
\qquad
\sigma_t=
\sqrt{
\frac{1}{N-1}\sum_{i=1}^{N}
\left(G_t^{(i)}-\mu_t\right)^2
}.
\label{eq:group_statistics}
\end{equation}

Based on the above, the group relative advantage is calculated:
\begin{equation}
\widehat A_t^{(i)}=
\frac{G_t^{(i)}-\mu_t}{\sigma_t+\delta},
\label{eq:group_advantage}
\end{equation}
where the advantage $\widehat A_t^{(i)}$ represents the future discounted returns of the sample relative to the same group.

\subsection{Target Clipping and Policy Optimization}

After the group relative advantage estimation, the actor policy is updated through multiple rounds of mini-batch optimization. The probability ratio of the new policy and old policy is defined as:
\begin{equation}
\rho_t^{(i)}(\theta)=
\frac{
\pi_\theta\left(a_t^{(i)}\mid s_t^{(i)}\right)
}{
\pi_{\theta_{\mathrm{old}}}
\left(a_t^{(i)}\mid s_t^{(i)}\right)
}.
\label{eq:policy_ratio}
\end{equation}

In practice, the normalized advantage is truncated to:
\begin{equation}
\widetilde A_t^{(i)}
=
\operatorname{clip}
\left(\widehat A_t^{(i)},-c,c\right),
\label{eq:advantage_clipping}
\end{equation}
where $c$ is the advantage truncation range. Then, the clipped surrogate objective is adopted to update the policy:
\begin{equation}
\begin{aligned}
\mathcal{L}(\theta)
={}&
\widehat{\mathbb{E}}_{\mathcal{D}}
\Bigl[
\min\Bigl\{
\rho_t^{(i)}(\theta)\widetilde A_t^{(i)},\\
&\quad
\operatorname{clip}
\bigl(\rho_t^{(i)}(\theta),1-\epsilon,1+\epsilon\bigr)
\widetilde A_t^{(i)}
\Bigr\}\\
&\quad
+\beta\,\overline{\mathcal{H}}_\theta
\bigl(s_t^{(i)}\bigr)
\Bigr],
\end{aligned}
\label{eq:policy_objective}
\end{equation}
where $\epsilon$ is the clipping ratio, and $\beta$ represents the entropy regularization coefficient.
$\overline{\mathcal H}_\theta$ represents the average policy entropy across all action dimensions. During the optimization process, the sampled log probabilities and advantages remain constant, while only the policy parameters are updated. The clipping target prevents excessive policy updates, and entropy regularization encourages exploration.

\section{Experiments}

\begin{figure*}[!t]
    \vspace{1.0em}
    \centering

    \begin{minipage}[t]{0.32\textwidth}
        \centering
        \includegraphics[width=\linewidth]
        {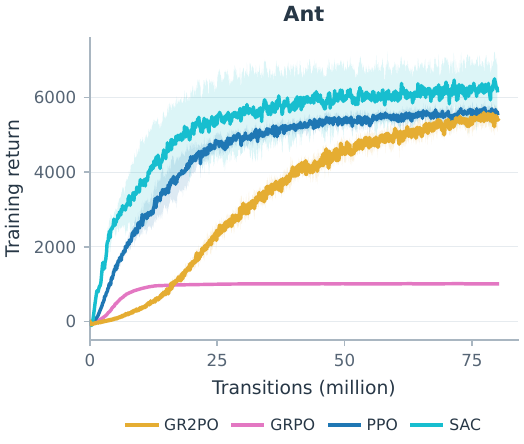}
        \par\vspace{2pt}
    \end{minipage}
    \hfill
    \begin{minipage}[t]{0.32\textwidth}
        \centering
        \includegraphics[width=\linewidth]
        {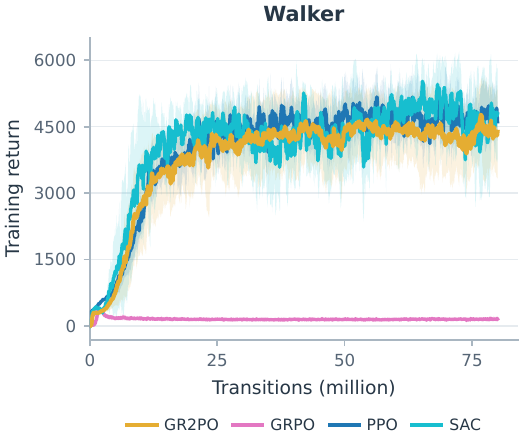}
        \par\vspace{2pt}
    \end{minipage}
    \hfill
    \begin{minipage}[t]{0.32\textwidth}
        \centering
        \includegraphics[width=\linewidth]
        {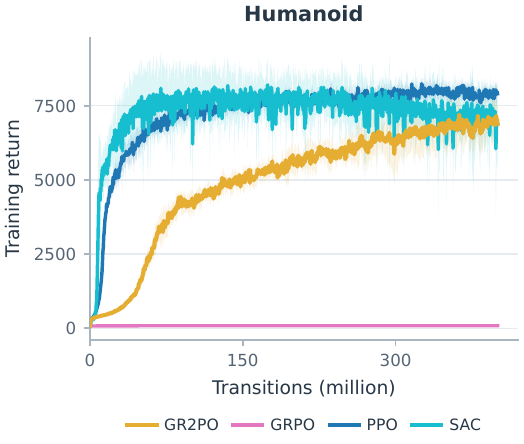}
        \par\vspace{2pt}
    \end{minipage}

    \caption{
        Training performance on three continuous control tasks. Training return is the average episode return over the last 100 completed training episodes. Lines and shaded regions show the mean and one standard deviation across five training seeds, respectively. Transitions are counted across all parallel environments.
    }
    \label{fig:training_curves}
\end{figure*}

\begin{table*}[!t]
    \centering
    \caption{
        Final training and evaluation returns
        on three continuous control tasks.
    }
    \label{tab:performance}

    \begingroup
    \small
    \setlength{\tabcolsep}{4pt}
    \renewcommand{\arraystretch}{1.15}

    \begin{tabular*}{\textwidth}
        {@{\extracolsep{\fill}}lcccccc@{}}
        \toprule
        &
        \multicolumn{2}{c}{Ant (80M)} &
        \multicolumn{2}{c}{Walker2d (80M)} &
        \multicolumn{2}{c}{Humanoid (400M)}
        \\
        Method
        & Training & Evaluation
        & Training & Evaluation
        & Training & Evaluation
        \\
        \midrule

        GR2PO
        & $\mathrm{5403.09}\pm\mathrm{344.14}$
        & $\mathrm{6335.25}\pm\mathrm{104.08}$
        & $\mathrm{4399.31}\pm\mathrm{1038.48}$
        & $\mathrm{5023.95}\pm\mathrm{737.66}$
        & $\mathrm{6896.20}\pm\mathrm{617.05}$
        & $\mathrm{8330.68}\pm\mathrm{759.20}$
        \\

        GRPO
        & $\mathrm{1006.23}\pm\mathrm{3.94}$
        & $\mathrm{1012.77}\pm\mathrm{5.86}$
        & $\mathrm{158.72}\pm\mathrm{48.96}$
        & $\mathrm{129.60}\pm\mathrm{40.87}$
        & $\mathrm{84.41}\pm\mathrm{0.46}$
        & $\mathrm{86.93}\pm\mathrm{2.37}$
        \\

        PPO
        & $\mathrm{5557.12}\pm\mathrm{54.23}$
        & $\mathrm{6609.79}\pm\mathrm{97.45}$
        & $\mathrm{4615.53}\pm\mathrm{433.92}$
        & $\mathrm{5964.93}\pm\mathrm{662.87}$
        & $\mathrm{7908.41}\pm\mathrm{96.86}$
        & $\mathrm{9484.00}\pm\mathrm{358.66}$
        \\

        SAC
        & $\mathrm{6225.22}\pm\mathrm{649.47}$
        & $\mathrm{6847.09}\pm\mathrm{549.63}$
        & $\mathrm{4794.23}\pm\mathrm{1014.92}$
        & $\mathrm{5294.46}\pm\mathrm{878.84}$
        & $\mathrm{6893.04}\pm\mathrm{1245.36}$
        & $\mathrm{7279.11}\pm\mathrm{1057.51}$
        \\

        \bottomrule
    \end{tabular*}

    \par\vspace{3pt}
    \begin{minipage}{\textwidth}
        \footnotesize
        Values are the mean and one sample standard deviation across five training seeds. Training denotes the final average return over the last 100 completed training episodes. Evaluation denotes the mean episode return of each final policy under a common evaluation scheme. 80M and 400M indicate total environment transitions.
    \end{minipage}
    \endgroup
    \vspace{-2.0em}
\end{table*}

\subsection{Experimental Setup}

We evaluate GR2PO on three continuous robot control tasks in the MuJoCo~\cite{mujoco} simulation platform, including Ant-v5, Walker2d-v5, and Humanoid-v5, which respectively involve quadruped robot movement, bipedal walking, and high-dimensional humanoid control. 
EnvPool~\cite{envpool} is adopted for parallel environment sampling.
All the training and evaluation run on a single NVIDIA RTX 4090 GPU with an AMD EPYC 9654 CPU. The maximum episode length for each environment is 1,000 steps. The episode can end early under the termination conditions. Training consumes 80 million environment transitions for both Ant and Walker2d, and 400 million for Humanoid, including all transitions from the parallel environments.

We compare GR2PO with the critic-free method GRPO~\cite{grpo} and the state-of-the-art actor-critic methods PPO~\cite{ppo} and SAC~\cite{sac}. GRPO baseline maintains the same configuration as GR2PO, and only uses immediate rewards instead of returns to calculate advantages. All training uses the same random seeds from 42 to 46. We present results as mean values and sample standard deviations. The training returns are calculated as the average cumulative return of the last 100 completed episodes, and the control performance is measured through an independent evaluation of the learned policy. The evaluation details, ablation configurations, and deployment test settings are introduced in the corresponding sections.

\subsection{Control Performance}

Fig.~\ref{fig:training_curves} shows the training curves for all methods. The immediate reward GRPO baseline remains at low training returns across all tasks. GR2PO can continuously improve the control policy, showing that taking future outcomes into consideration plays a crucial role in effective learning. Compared with actor-critic baselines, GR2PO achieves comparable final training returns across all environments. These results indicate that GR2PO can obtain effective policies without the need to learn a critic. 

To test the final control performance, we perform independent evaluations for the learned policies. As shown in Table~\ref{tab:performance}, GR2PO achieves significantly higher final training and evaluation returns than the immediate reward GRPO baseline in all tasks. This supports the effectiveness of incorporating future rewards into the group relative advantage estimation. In the deterministic policy evaluation, the mean returns of GR2PO on Ant, Walker2d, and Humanoid reach 6335.25, 5023.95, and 8330.68, respectively. Compared with the actor-critic baselines, GR2PO performs competitively in evaluation returns on Ant and Walker2d, outperforms SAC on Humanoid but underperforms PPO. Furthermore, the standard deviation of the evaluation return for GR2PO across all tasks is lower than SAC. This indicates improved stability across training seeds compared to SAC. Overall, GR2PO can learn effective continuous control policies without learning a critic and achieves competitive control performance on the tested tasks.

\subsection{Training Efficiency}

\begin{figure*}[!t]
    \vspace{1.0em}
    \centering
    \envpanel{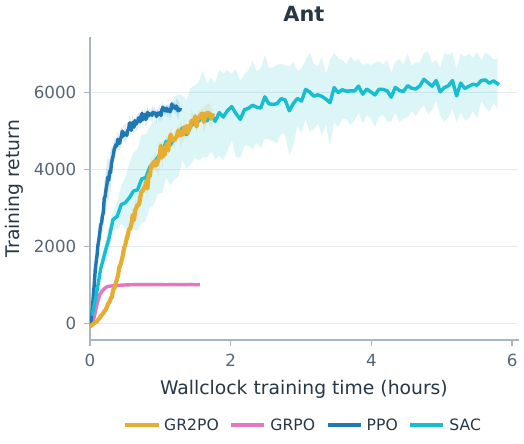}
    \hfill
    \envpanel{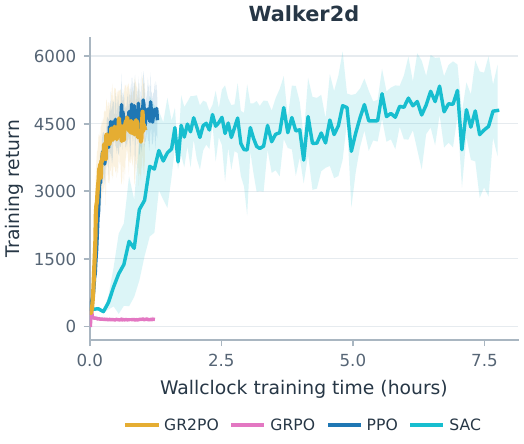}
    \hfill
    \envpanel{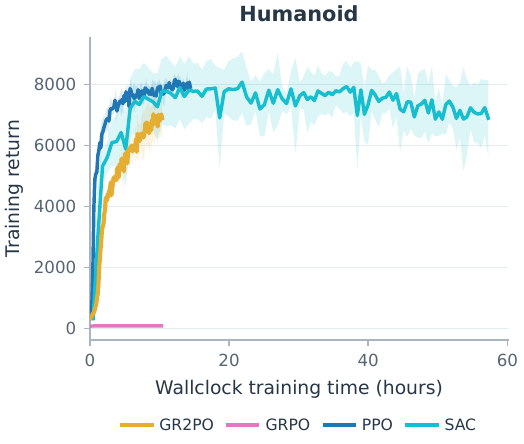}

    \caption{
        Training return versus wall-clock training time. Runs are aligned at matched training progress. We average training time and return across five training seeds. Lines and shaded regions show the mean and one standard deviation of return.
    }
    \label{fig:wallclock_curves}
    \vspace{-2.0em}
\end{figure*}

\begin{figure}[!t]
    \vspace{1.0em}
    \centering
    \includegraphics[width=\columnwidth]
    {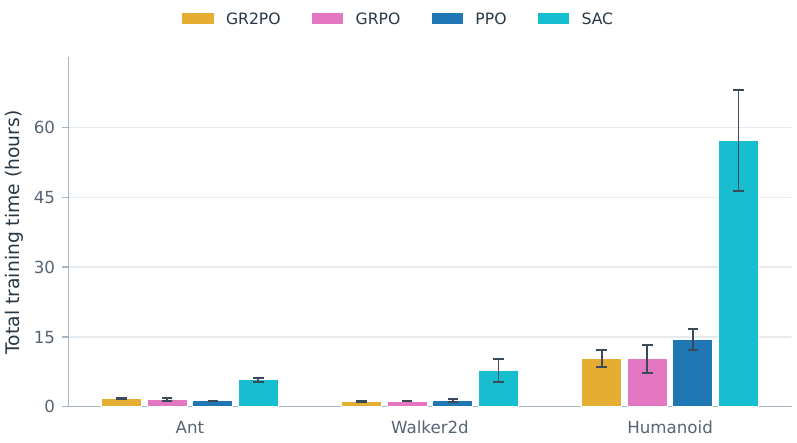}

    \caption{
        Total wall-clock training time at fixed transition budgets: 80M for Ant and Walker2d, and 400M for Humanoid.
        Bars and error bars indicate the mean and
        sample standard deviation across five training seeds, respectively. Lower values indicate shorter training times for the specified transition budget.
    }
    \label{fig:training_duration}
    \vspace{-3.0em}
\end{figure}

To evaluate the training efficiency of the proposed framework, We combine return-time training curve and the total wall-clock training time for analysis. The former shows how the policy performance changes with the training time, while the latter measures the time required by each method to complete the same interaction budget.
As shown in Fig.~\ref{fig:training_duration}, compared to SAC, GR2PO completes the training with a shorter total wall-clock time in all three environments, and achieved competitive control performance. Compared with PPO, GR2PO requires less total training time on Walker2d and Humanoid environments. Therefore, GR2PO has demonstrated superior training speed. As shown in Fig.~\ref{fig:wallclock_curves}, PPO has faster performance improvement in the early learning phase of some tasks. Therefore, the speed to complete a fixed transition budget needs to be distinguished from the speed to achieve a specific performance. The results show that GR2PO provides a competitive trade-off between control performance and training costs. The critic-free setting reduces the computational burden involved in value network training, but the actual efficiency is still influenced by both the sampling scale and the configuration of policy updates.

\subsection{Ablation Studies}

To analyze the key design of GR2PO, we evaluate the effects of parallel group size, return estimation method, and clipping ratio parameters on learning performance.

\subsubsection{Group Size Study}

\begin{figure*}[!t]
    \vspace{1.0em}
    \centering
    \envpanel{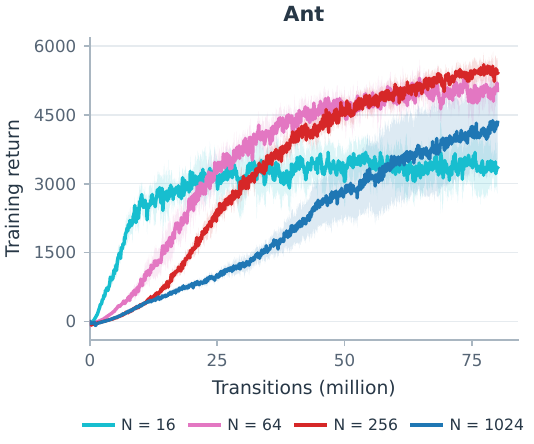}
    \hfill
    \envpanel{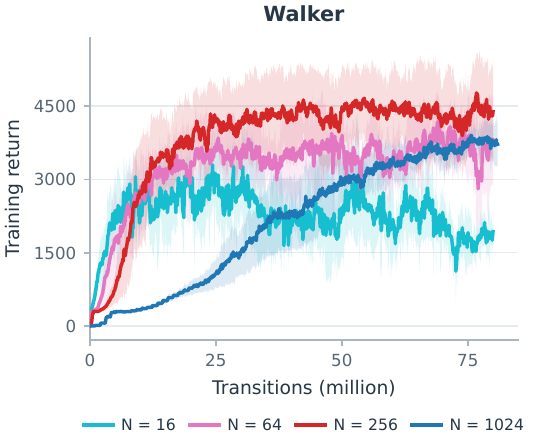}
    \hfill
    \envpanel{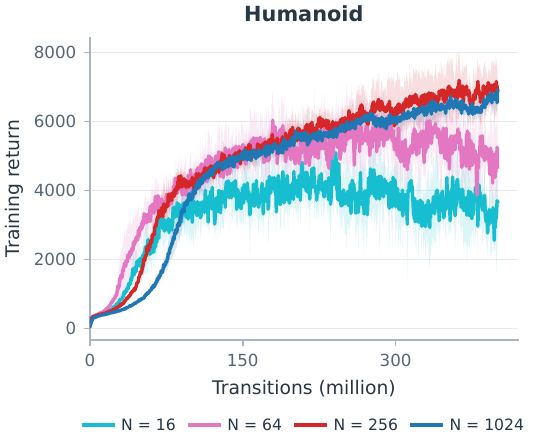}

    \caption{
        Effect of group size on GR2PO.
        Within each task, all runs use the same fixed total transition budget, with update epochs and minibatch size adjusted for the corresponding parallel configuration.
        Lines and shaded regions indicate the mean and one standard deviation across five training seeds.
    }
    \label{fig:group_size}
\end{figure*}

We study the influence of group size on the training process of GR2PO, where group size refers to the number of parallel environments. All the configurations take same total transition budgets, and the minibatch size and the update epochs are adjusted correspondingly. As shown in Fig.~\ref{fig:group_size}, the group size affects not only the final return but also the speed of the performance improvement.
A small group size converges rapidly in the early training, but has a lower upper bound. A larger group size may require more interaction to demonstrate its advantages, and it may not necessarily lead to the highest return within a limited budget. The medium group size achieves the highest final return. 

The results demonstrate that increasing group size provides more samples to compare across environments. It will also change the sampling quality and the number of updates. Therefore, the selection of group size needs to take into account the relationship between the advantage estimation and the policy update. This ablation study reflects the overall performance under different parallel sampling and update configurations, showing that appropriately matching the group size and update settings is crucial for maximizing the performance of GR2PO.

\begin{figure*}[!t]
    \centering
    \envpanel{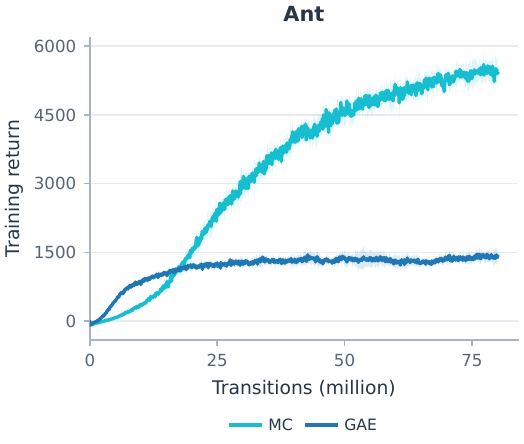}
    \hfill
    \envpanel{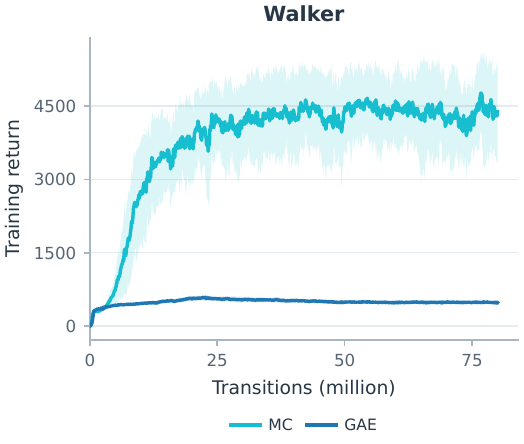}
    \hfill
    \envpanel{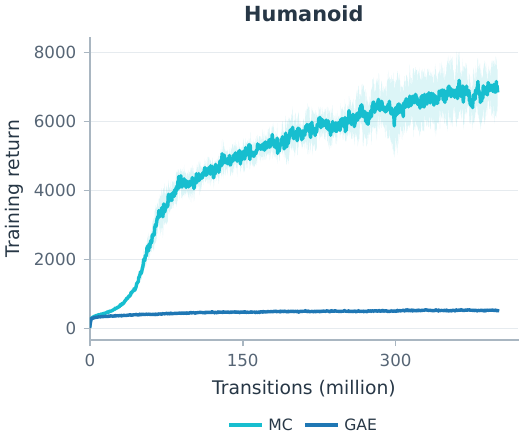}

    \caption{
        Effect of return estimation methods on GR2PO. We compare Monte Carlo (MC) and a Generalized Advantage Estimation (GAE) variant using a group statistics baseline. Lines and shaded regions indicate the mean and one standard deviation across five training seeds.
    }
    \label{fig:return_estimation}
\end{figure*}

\begin{figure*}[!t]
    \centering
    \envpanel{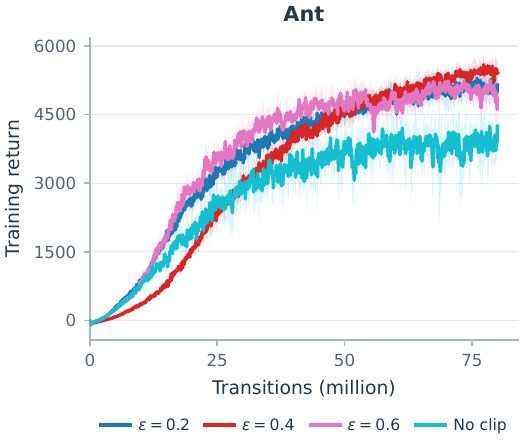}
    \hfill
    \envpanel{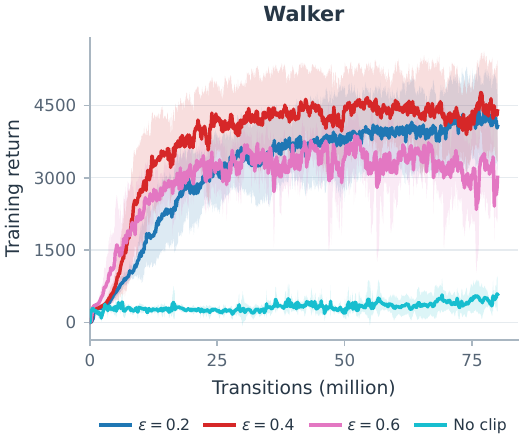}
    \hfill
    \envpanel{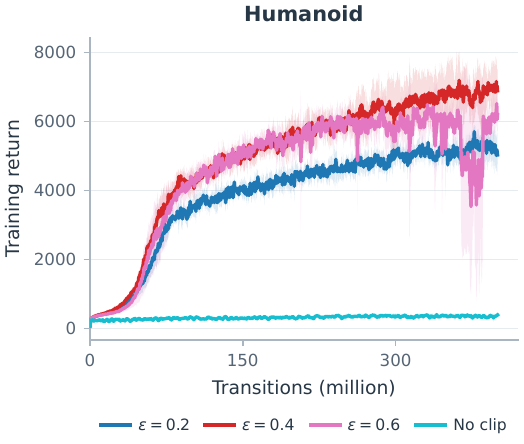}

    \caption{
        Effect of the target ratio clipping parameters on GR2PO. The no-clipping configuration removes the target clipping mechanism while keeping advantage truncation and KL-based early stopping. Lines and shaded regions indicate the mean and one standard deviation across five training seeds.
    }
    \label{fig:clipping}
    \vspace{-2.0em}
\end{figure*}

\subsubsection{Return Estimation Methods}

To analyze the effects of return estimation methods, we compare Monte Carlo (MC)~\cite{sutton2018} approach and a Generalized Advantage Estimation (GAE)~\cite{gae} variant. For the critic-free setting, the GAE~\cite{gae} variant uses the average return across all time steps as the baseline to build the TD error. As shown in Fig.~\ref{fig:return_estimation}, MC supports consistent policy improvement, while the GAE~\cite{gae} variant fails to learn effective policies and only demonstrates a similar level to the immediate reward baselines. MC directly uses the future returns in the sampling trajectories, allowing the relative evaluation of the current action to be incorporated into the long-term results, without relying on the learned value function. Therefore, this ablation study reflects the combined influence of the estimation method and the normalization method, which supports MC to serve as the default return estimation method for GR2PO.

\subsubsection{Clipping Ratio Study}

We further analyze the effects of update constraints on GR2PO by comparing different target clipping ratio parameters and the no-clipping setting. As shown in Fig.~\ref{fig:clipping}, moderate target clipping achieves better training performance across all environments, while excessive or insufficient clipping may both affect the learning outcome. 
We analyze the reason is that a smaller value has limited the contribution of samples to the policy update and possibly slowed down the performance improvement. As the clipping ratio increases, the policy update adjusts more flexibly but the restrictions on the change in probability ratio also weaken accordingly. The experiments of no-clipping setting further indicates that relying solely on the group relative advantages is not sufficient to achieve the best training performance. As a result, the policy update surrogate objective needs appropriate clipping. This ablation study indicates that the return estimation and update constraints play different roles in GR2PO. The former determines the temporal information contained in the action evaluation, while the latter regulates how these evaluations applied to policy optimization. Therefore, appropriate clipping is an crucial component of this critic-free framework.

\subsection{Policy Deployment on Jetson TX2}

\begin{table*}[!t]
    \vspace{1.0em}
    \centering
    \caption{Policy deployment on NVIDIA Jetson TX2.}
    \label{tab:tx2_deployment}

    \begingroup
    \small
    \setlength{\tabcolsep}{4pt}
    \renewcommand{\arraystretch}{1.15}

    \begin{tabular*}{\textwidth}
        {@{\extracolsep{\fill}}llrrcc@{}}
        \toprule
        Environment
        & Method
        & Actor params.
        & Size (MiB)
        & CPU latency (ms)
        & CUDA latency (ms)
        \\
        \midrule

        Ant
        & GR2PO & 126,856 & 0.488
        & $0.731 \pm 0.012$
        & $1.357 \pm 0.030$
        \\
        & PPO & 126,856 & 0.488
        & $0.763 \pm 0.075$
        & $1.353 \pm 0.029$
        \\
        & SAC & 126,856 & 0.488
        & $0.782 \pm 0.149$
        & $1.353 \pm 0.015$
        \\

        \addlinespace[4pt]

        Walker2d
        & GR2PO & 104,070 & 0.401
        & $0.689 \pm 0.003$
        & $1.299 \pm 0.060$
        \\
        & PPO & 104,070 & 0.401
        & $0.684 \pm 0.003$
        & $1.282 \pm 0.028$
        \\
        & SAC & 104,070 & 0.401
        & $0.688 \pm 0.002$
        & $1.278 \pm 0.014$
        \\

        \addlinespace[4pt]

        Humanoid
        & GR2PO & 607,761 & 2.325
        & $2.645 \pm 0.029$
        & $1.519 \pm 0.018$
        \\
        & PPO & 607,761 & 2.325
        & $2.630 \pm 0.044$
        & $1.519 \pm 0.046$
        \\
        & SAC & 607,761 & 2.325
        & $2.572 \pm 0.008$
        & $1.509 \pm 0.008$
        \\

        \bottomrule
    \end{tabular*}

    \par\vspace{3pt}
    \begin{minipage}{\textwidth}
        \footnotesize
        Deterministic FP32 policy inference with batch size 1. Latency is measured end-to-end from a CPU observation to a CPU action, including preprocessing and applicable device transfers and synchronization. Values are the mean and one sample standard deviation over 10 timing blocks with 2000 calls per block. Size includes the exported actor and observation normalization state, excluding critic networks.
    \end{minipage}

    \endgroup
\end{table*}

\begin{figure}[t]
    \vspace{1.0em}
    \centering
    \includegraphics[width=\columnwidth]{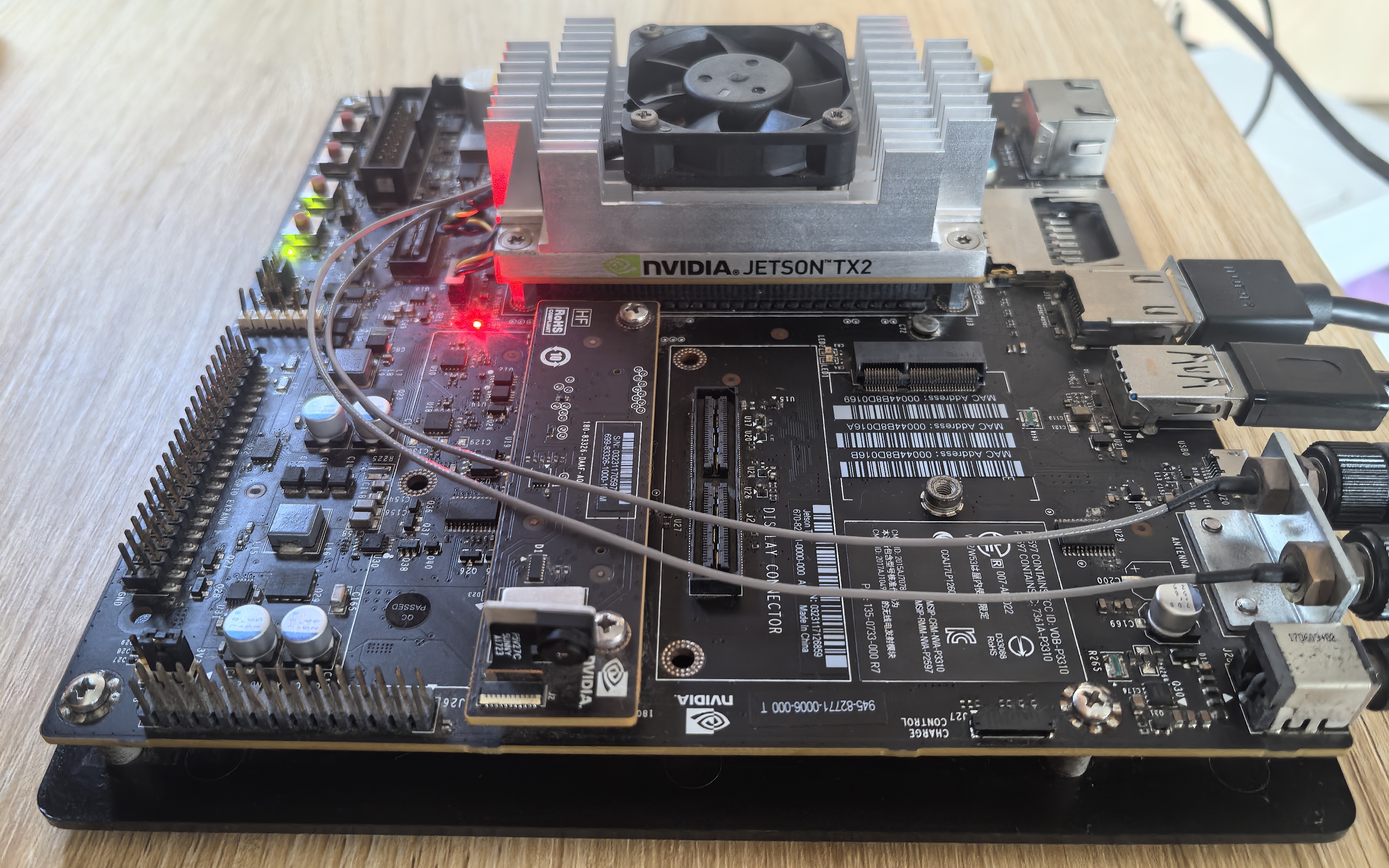}
    \caption{NVIDIA Jetson TX2 platform used for policy inference tests. The trained actor and observation normalization statistics are deployed on this platform. Inference latency is measured on both CPU and GPU.}
    \label{fig:jetsontx2}
    \vspace{-2.0em}
\end{figure}

To assess the deployability of the policies trained by GR2PO on resource-constrained hardware, we evaluate policy inference on the NVIDIA Jetson TX2 platform shown in Fig.~\ref{fig:jetsontx2}. The evaluation continuous control tasks are the same as the training stage, including Ant, Walker2d and Humanoid.
We test separately on the CPU and the CUDA backend. We compare GR2PO, PPO, and SAC in terms of inference latency, parameter count, and model size. Applicable device transfers and synchronization costs are included, excluding environmental simulation and sensor or actuator delays. For a consistent comparison, only the trained actors and the associated observation normalization statistics are deployed on the TX2 platform, without the critic used in the training stage. Inference is performed with a batch size of one, corresponding to generating an action from a single observation at each control step. All the models use deterministic action output at FP32 precision.

As shown in Table~\ref{tab:tx2_deployment}, the policies trained by GR2PO supports low-latency inference on the TX2 platform. The model size and the actor parameter counts are same in each task. The use of a critic during training does not directly determine the deployed parameter count, since PPO and SAC also require only their actors for action generation. In this case, GR2PO simplifies training process due to the critic-free design. The inference costs depend more on the policy network architecture and the observation dimensions. The CPU and CUDA latency measure policy inference under different computational configurations. The results demonstrate that our trained model achieves low-latency inference on a real-world edge device. Moreover, GPU devices can not necessarily result in lower latency for lightweight policy networks with single-observation inputs.

This experiments provide hardware verification for GR2PO beyond the simulation control performance, showing trained policies can run on a physical edge device. Return estimation and group normalization are only used during training process and introduce no additional computation during deployment. However, these tests measure policy inference instead of the complete control loop. Therefore, the results of the TX2 deployment support the feasibility of edge inference, while closed-loop performance remains to be validated.

\section{Conclusion}

This paper proposes Group Relative Return Policy Optimization (GR2PO), a critic-free reinforcement learning framework for continuous robot control. GR2PO constructs relative advantages via return estimation and group normalization, and uses clipped targets to update the actor policy without learning a critic. The simulation experiments demonstrate that GR2PO significantly outperforms critic-free baselines with immediate rewards. Moreover, it achieves competitive control performance and training efficiency on multiple tasks. Through further ablation studies, we analyze the effects of parallel group size, return estimation methods, and target clipping ratio on learning performance. Inference tests on NVIDIA Jetson TX2 demonstrate the feasibility of deploying the learned policies on this edge platform. We will release all code and experimental configurations.

This study still has several limitations. The experiment mainly covers three motion control tasks. Deployment on edge platforms involves only policy inference, without verifying closed-loop control performance. In the future, we will expand GR2PO to more robot tasks, further study the applicability of group relative return, and validate its performance in real-world closed-loop control.


%%%%%%%%%%%%%%%%%%%%%%%%%%%%%%%%%%%%%%%%%%%%%%%%%%%%%%%%%%%%%%%%%%%%%%%%%%%%%%%%

%%%%%%%%%%%%%%%%%%%%%%%%%%%%%%%%%%%%%%%%%%%%%%%%%%%%%%%%%%%%%%%%%%%%%%%%%%%%%%%%

%%%%%%%%%%%%%%%%%%%%%%%%%%%%%%%%%%%%%%%%%%%%%%%%%%%%%%%%%%%%%%%%%%%%%%%%%%%%%%%%

%\newpage

\end{document}